\documentclass[conference]{IEEEtran}
\usepackage{cite}

\usepackage{hyperref}

\newcommand{\etal}{\textit{et al.}}

\usepackage{amsmath,amssymb,amsfonts}
\usepackage{algorithmic, algorithm}
\usepackage{graphicx}
\usepackage{textcomp}
\usepackage{xcolor}
\def\BibTeX{{\rm B\kern-.05em{\sc i\kern-.025em b}\kern-.08em
    T\kern-.1667em\lower.7ex\hbox{E}\kern-.125emX}}
\begin{document}

\title{Accelerating the Learning of TAMER with Counterfactual Explanations}

\author{\IEEEauthorblockN{Jakob Karalus}
\IEEEauthorblockA{\textit{Institute of Artificial Intelligence} \\
\textit{Ulm University}\\
Ulm, Germany \\
jakob.karalus@uni-ulm.de}
\and
\IEEEauthorblockN{Felix Lindner}
\IEEEauthorblockA{\textit{Institute of Artificial Intelligence} \\
\textit{Ulm University}\\
Ulm, Germany \\
felix.lindner@uni-ulm.de}
}

\maketitle

\begin{abstract}
The capability to interactively learn from human feedback would enable agents in new settings. For example, even novice users could train service robots in new tasks naturally and interactively.  
Human-in-the-loop Reinforcement Learning (HRL) combines human feedback and Reinforcement Learning (RL) techniques. 
State-of-the-art interactive learning techniques suffer from slow learning speed, thus leading to a frustrating experience for the human.
We approach this problem by extending the HRL framework TAMER for evaluative feedback with the possibility to enhance human feedback with two different types of counterfactual explanations (action and state based). 
We experimentally show that our extensions improve the speed of learning.
\end{abstract}

\begin{IEEEkeywords}
Human-in-the-loop Reinforcement Learning, Counterfactuals, Explainability
\end{IEEEkeywords}

\section{Introduction}
\label{intro}

Classical Machine Learning approaches like Supervised Learning or Reinforcement Learning can solve complex tasks with sophisticated architectures, but they allow for little to no interaction possibilities with the human throughout their training. The field of Human-in-the-loop Reinforcement Learning (HRL) aims to integrate the human into the learning process, allowing the agent to learn directly from the human.
This integration of humans into the training process allows non-experts to train or modify an agent's behavior naturally. 


While integrating humans into the learning process opens up new possibilities, doing so also introduces new 
challenges. 
In this work, we address two of them: the integration of an enhanced interface for evaluative feedback and the learning speed of the algorithm. When teaching an agent, humans naturally want to enrich their feedback with explanations, and a too simplistic interface can lead to frustration on the human side, see \cite{HumanExperienceIRL,UserFeedbackMachineLearning}. In most previous work, humans can only give simple binary feedback \cite{Knox2008TAMERTA} or control a single dimension \cite{Celemin2015COACHLC}.
Moreover, human feedback is more costly than automated feedback from the environment or a simulator. 
State-of-the-art Reinforcement Learning needs millions of episodes with billions of frames (see \cite{SampleEfficentRL}). It is insufficient to use the same algorithms and replace the environmental reward with the human reward since it is not feasible for a human to watch the agent for weeks or months of training. In addition to the time constraint, if progress is slow, it can lead to early frustration for the human if they see no impact of their feedback in training. Therefore, the algorithms' learning speed is crucial for a positive user experience. 

Since the flexible interfaces and learning speed are both crucial for the experience of the human, we want to investigate with this work if the inclusion of counterfactuals in TAMER can be beneficial for the learning speed. Counterfactuals are methods of explanations to describe a causal link between two possible outcomes in the form of:  ``If event B rather than A had happened, then outcome Y rather than X would have happened''. Since counterfactuals contain more information (due to the second outcome, see Section \ref{sec:Counterfactuals}), we believe they could be a solution to enhance the learning speed. In concrete, we focus on optional upward-directed counterfactuals, which should speed up learning and give humans more flexibility in their feedback when working with TAMER. We show that this type of feedback can be easily added to one of the most prominent methods for evaluative feedback, TAMER.

Source code and evaluations are available under: \url{https://anonymous.4open.science/r/InteractiveRL-25A1}.

\section{Related Work}
One of the first proposed frameworks for HRL with evaluative feedback is the TAMER framework \cite{Knox2008TAMERTA}, which we use as a basis for our work (see chapter \ref{sec:Background and Methods}). Throughout this work, we use the standard extension DeepTAMER \cite{Warnell2018DeepTI}, which enables the usage of Deep Neural Nets as a function approximator.
Multiple extensions of TAMER tackle the problem of combining human feedback with environmental feedback. TAMER+RL \cite{Knox2010CombiningMF} and continuous TAMER+RL \cite{Knox2012ReinforcementLF} both re-introduce the environmental reward and enable learning from multiple feedback sources. A common trend is introducing Reinforcement Learning frameworks into TAMER: Arakawa \etal  \cite{Arakawa2018DQNTAMERHR} combine it with Deep-Q learning, and Vien \etal \cite{Vien2012LearningVH} use an actor-critic setting to enable the usage in continuous-actions spaces. 
While all these approaches enable the usage of TAMER in additional types of environments, their assumption and interface for human feedback stay the same. Therefore our work would also be applicable to each of them.

COACH \cite{COACHMcGlasham} assumes that the feedback given by the human is advantage-based ("How much better is the current policy than the previous one?"), which allows to learn a policy directly. Since the significant difference between COACH and TAMER are the different assumptions about human feedback (forward-looking versus advantage-based) both variants are targeted toward different types of feedback. While our work is transferable to COACH, the crucially different assumption about how humans formulate feedback makes it hard to compare both methods fairly on a technical basis. At the same time, it is  still open which assumptions are preferred by humans. Therefore we refrain from an in-depth comparison of different both approaches. 

In contrast to TAMER's reward-shaping approach, a different approach to HRL is policy-shaping \cite{NIPS2013_e034fb6b}. Policy-shaping uses the feedback directly as labels for the corresponding policy instead of interpreting it as a reward signal. A prominent variant of this is the work of Celemin \etal \cite{Celemin2015COACHLC} for their COACH framework (a different COACH than the above mentioned), which can learn a policy from corrections. The concept of corrections 
bears similarities to our counterfactual actions. We refrain from a direct evaluation of counterfactuals in this framework due to the incompatibilities in the action spaces (their correction are on continuous actions, while ours are currently on discrete actions only). 

A different avenue for feedback in HRL has been preference-based methods, \cite{Akrour2011PreferenceBasedPL} where humans can give preference for one whole trajectory over another. 
This type of feedback requires the generation of trajectories, which can limit the application for inherently safe applications or simulated environments. Nevertheless, while our work is not directly applicable, the concepts should be easily adaptable for preference-based methods. 

Action Advising \cite{Torrey2013TeachingOA} is a popular feedback method for HRL, in which the human advises the agent on which actions it should perform next. Conceptually, advising is similar to our counterfactual actions, with the difference that advice is given beforehand instead of afterward. Moreover, a critical difference between Action Advising and our counterfactual actions is the contrastive nature of counterfactuals (see Section \ref{sec:Using the contrastive nature of counterfactuals}). Since our work is motivated by allowing more flexible feedback from humans rather than benchmarking different interfaces for feedback in HRL, we refrain from a technical comparison and refer to \cite{ijcai2021-interaction-considerations}. 


Guan \etal \cite{Guan2020ExplanationAF} integrate human explanation into the HRL by using saliency-based attention annotation combined with Deep Convolutions Networks. This method showed that the inclusion of explanations in HRL could lead to better performance. However, this method can only be applied in specific cases due to Saliency Maps and Deep Neural Nets. In contrast, our current proposed solution is primarily independent of the input shape and the approximator's chosen architecture. 

While counterfactuals have not been used in HRL, other work has successfully used counterfactuals in different areas. Lu \etal \cite{Lu2020SampleEfficientRL} have shown that an integration of counterfactuals, through the usage of Structural Causal Models (SCM), in Reinforcement Learning (without human feedback) can be beneficial to  label efficiency. 
Work by Jin \etal \cite{Jin2018RegretMF,Buesing2019WouldaCS} used counterfactuals in partial observable settings to increase the learning speed, and robustness of their Reinforcement Learning approaches.  Foerster \etal \cite{Foerster2018CounterfactualMP} use a counterfactual loss in a multi-agent reinforcement setting to increase their performance. Menglin \etal \cite{SARSACounterfactual} use counterfactual experiences to solve inefficiencies with exploration in the early stages of SARSA with counterfactuals. 

The concept of counterfactuals has been included in different applications with success. With this work, we 
formulate a general method for the inclusion of counterfactuals in HRL.

\section{Background and Methods}
\label{sec:Background and Methods}

This Section explains the relevant background and highlights our extension of the existing method. First, we will explain TAMER, introduce counterfactuals, and investigate the convergence of TAMER from a Reinforcement Learning perspective. Afterward we use this perspective as motivation to introduce counterfactuals in TAMER. 

\subsection{The TAMER Framework}
Human-in-the-loop Reinforcement Learning (HRL) allows the agent to learn from human feedback directly and incrementally, even in environments with no other reward signal. 
A major difference of human feedback is the potential inconsistency and sub-optimality of human feedback. Humans can give feedback at various stages throughout the training, but the agent cannot depend on a fixed and reliable reward (unlike the standard environmental reward). HRL is also advantageous in very complex environments where a specification of a solid reward function is challenging since even slight misspecifications of the reward function can lead to unexpected side-effects in actual use cases \cite{dulac2019challenges}.

In this work, we use TAMER, a framework for evaluative feedback. In this scenario, after the agents performed an action $a$ in a specific state $s$, the human has the opportunity to give his evaluative feedback $\mathit{f}$ towards the agent. No additional reward function is given and the agent has to learn a policy from the human reward. In most work the range of the feedback is restricted towards discrete feedback of $-1$ and $+1$.



TAMER \cite{Knox2008TAMERTA} allows to learn a policy with human evaluative feedback by assuming that the human already considers the long-term solution in his feedback. This assumption lets us greedily learn a function that maps states 
to their excepted feedback: ($H: S\times A \rightarrow F$). In TAMER, mapping $H$ is learned immediately with a single \textit{state/action} pair when feedback is given. This learned function of the human reward ($H$) is used to greedily select actions with the highest expected human reward at each time-step, i.e., $argmax_a(H(s,a))$. Therefore, TAMER is conceptually similar to Q-Learning with the crucial difference that it assumes long-term feedback and therefore does not have the credit assignment problem. TAMER does not use any environmental reward, methodical exploration, or longer-term planning mechanism.


DeepTAMER \cite{Warnell2018DeepTI} extends TAMER by using Deep Neural Networks as the approximator for the $H$ function. The network architecture consists of shared base layers and individual fully-connected heads for each action. Additionally, a replay buffer (which stores past feedback/states) is used to stabilize training. In our work, we will roughly follow the architecture of DeepTAMER, with minor modifications due to different state space and action space sizes. 


\subsection{Counterfactuals}
\label{sec:Counterfactuals}
Counterfactual explanations describe a causal connection between two possible events in the form: ``If event B rather than A had happened, then outcome Y rather than X would have happened'' \cite{byrne2019counterfactuals}. Counterfactuals thus refer to an event $A$ that  actually has happened and caused some outcome \emph{fact} $X$, and to the \textit{contrast} $B$, which would have resulted in a different outcome (the \emph{foil}) $Y$ \cite{miller2019explanation}. It is important to note that a counterfactual does not only provide a second data point---the \textit{contrast-foil pair}---, but also the \textit{contrast relationship} between the fact and the foil is an equally important property. 
We will make use of both these properties, that is, we treat the foil as an additional example and exploit the contrastive relationship between fact and foil to achieve  faster learning.


As multiple studies \cite{CounterfactualThinking,CounterfactualThinking2} show, it is easier for most people to envision a world ``better-off'' than a world ``worse-off''. Additionally, positively-directed counterfactuals are usually of higher quality \cite{byrne2019counterfactuals}. 
We therefore allow humans to state counterfactual explanations of negative feedback to an observed event $\langle state, action\rangle$. The counterfactuals can take the form ``If $a$ had been performed in state $s_{\mathit{cf}}$ rather than in state $s$, then my feedback would have been positive rather than negative'' and ``If action $a_{\mathit{cf}}$ had been performed $s$ rather than action $a$, then my feedback would have been positive rather than negative.''
This allows the agent to learn how to get from a state with bad feedback into a state with good feedback. 

\begin{figure}[t] 
     \centering
    \includegraphics[width=0.49\textwidth]{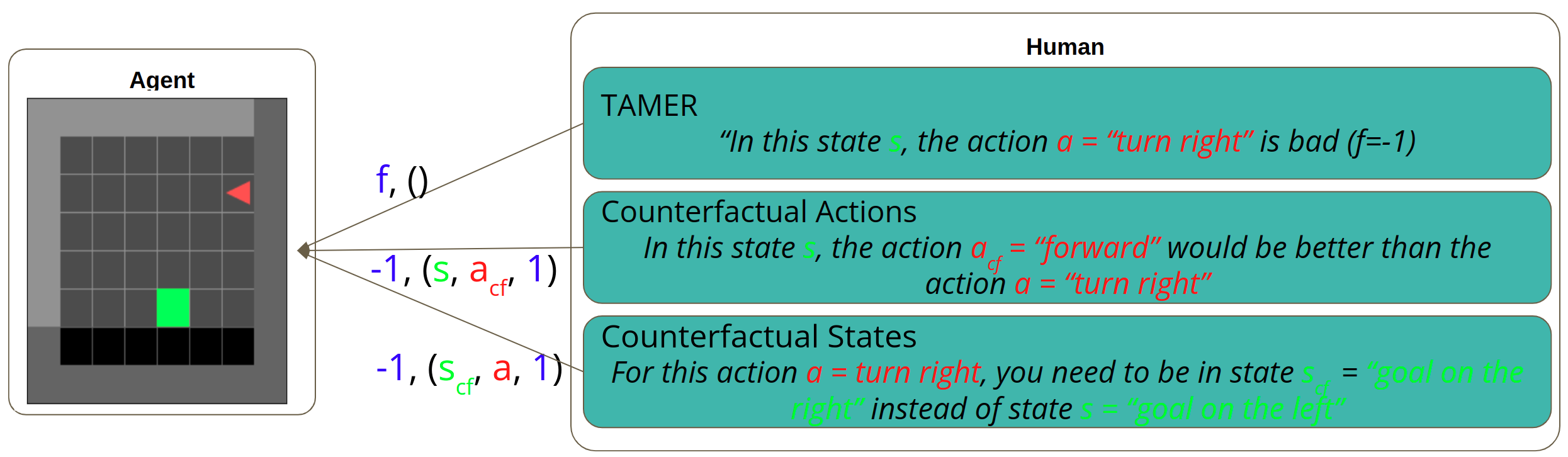}
\caption{Left: Visualization of the MiniGrid environment with the goal as a green square. The highlighted area indicates the current viewpoint of the agent. Right: Verbalization of the various feedback mechanism in natural language.}
\label{fig:environment}
\end{figure}
\subsection{The convergence of Reinforcement Learning algorithms}
\label{sec:The convergence of Reinforcement Learning algorithms}
In HRL,  learning speed is 
particularly essential because humans are asked to give feedback to the learner. This is a tedious task we naturally want to minimize. An agent which learns too slowly or shows no progress can limit the motivation of the human to give further feedback \cite{HumanFeedbackMotivation}. 
In Reinforcement Learning, the convergence difference between an expert policy $\pi^{*}$ and a sub-optimal $\pi$ can be described by the distance between the optimal distribution of visited states and the current distribution (also known as \textit{performance difference lemma} due to Kakade \cite{Kakade2003OnTS}):
\begin{equation}
    V^{\pi^{*}} - V^{\pi} = \sum_{s} \pi^{*}(s)\sum_{a}(\pi^{*}(a\mid s)- \pi(a\mid s))Q^{\pi}(s,a)
    \label{eq:performance difference lemma}
\end{equation}
The optimal policy is denoted by $\pi^{*}$, $\pi^{*}(s)$ is the visited state distribution of $\pi^{*}$, and $\pi^{*}(a\mid s)$ is the action distribution of the optimal policy. 
Accordingly, there will always be a difference in performance as long as there is a substantial difference in the visited states. While the original lemma was formulated for classical Q-learning, we apply the same concept to TAMER since its replacement of $Q$ with $H$ can be seen as identical in function and convergence. 
In a setting with an ample state/action space, under a random policy $\pi$, the distribution of visited states will drastically differ from the expert distribution. This leads to a ``chicken/egg'' problem, where to learn quickly, an agent has to reach states similar to the states the expert would reach, but doing that is essentially the goal of learning. 

The previously introduced concept of positively-directed counterfactuals comes directly from the expert distribution since it indicates which part of the current $\langle state, action \rangle$ pair should have been different.
This should allow the agent to reduce the difference $V^{\pi^{*}} - V^{\pi}$, which leads to faster learning. 
To experimentally test this hypothesis, we introduce two different types of positively-directed counterfactuals into the TAMER framework and confirm that their inclusions lead to faster learning. 
Besides the above-discussed human reasons for ``better-off'' counterfactuals, the \textit{performance difference} thus offers an additional viewpoint favoring our focus on positively-directed counterfactuals. 

\subsection{Using counterfactuals in TAMER}

To introduce counterfactuals into TAMER, we have to change the \textit{fact} of our current $\langle state, action\rangle$ pair, so that the new \textit{contrast} would receive a positive reward. To achieve this contrast, we modify either the state or the action of our fact. We call the modified element \textit{counterfactual state} or \textit{counterfactual action}. In the following we will denote counterfactual actions or states with the subscript $s_{\mathit{cf}}$ and $a_{\mathit{cf}}$. When asking the human for feedback with $gather\_\mathit{f}eedback(s,a)$ the subscripts $s_{t-1}$, $a_{t-1}$ indicate the last state or action performed (since the human gives their feedback after an action is performed, so the agent is already in a new state).

A counterfactual state provides a state where the action would be a good one. The feedback $gather\_\mathit{f}eedback(s_{t-1}, a_{t-1}) = (-1, \langle 1, s_{\mathit{cf}}, a \rangle$) can be read as \textit{``Action $a$ would be good if you were in state $s_{\mathit{cf}}$ instead of $s$''}. 
In case of counterfactual actions, $gather\_\mathit{f}eedback(s_{t-1}, a_{t-1}) = (-1, \langle 1, s, a_{\mathit{cf}} \rangle$) can be described in natural language with: \textit{``In state $s$, action $a_{\mathit{cf}}$ would be better than action $a$''}. Both types of counterfactuals are exemplified in Figure \ref{fig:environment}.

\begin{algorithm}[t]
\begin{algorithmic}
\REQUIRE $H$ function approximator; total episodes $N_e$; maximum number of steps per episode $N_s$
\FOR{e = 1 \textbf{to} $N_e$} 
\FOR{t = 1 \textbf{to} $N_s$} 
\STATE $s_{t}$ $ \leftarrow $ observe state
\STATE $f, \langle f_{\mathit{cf}}, s_{\mathit{cf}}, a_{\mathit{cf}} \rangle \leftarrow$ gather feedback ($s_{t-1}$, $a_{t-1})$
\IF{$(f_{\mathit{cf}} )$}
\STATE update $H$ with $L(s_{t-1}, a_{t-1}, f,  s_{\mathit{cf}}, a_{\mathit{cf}}, f_{\mathit{cf}})$
\ELSE
\STATE update $H$ with $\langle f, s_{t-1}, a_{t-1}\rangle$
\ENDIF
\STATE $a_t$ $\leftarrow$ $argmax_a(H(s,a))$
\STATE perform action $a_t$
\ENDFOR
\ENDFOR
\end{algorithmic}
\caption{TAMER with Counterfactual Feedback}
\label{algo:tamer}
\end{algorithm}

To include these counterfactuals in TAMER, we extend the human feedback function to return an optional triple $\langle \mathit{f}_{\mathit{cf}}, s_{\mathit{cf}}, a_{\mathit{cf}} \rangle$. This triple contains the counterfactual feedback $\mathit{f}_{\mathit{cf}}$ and either $s_{\mathit{cf}}$ or $a_{\mathit{cf}}$ contains a counterfactual version of the original input, depending on the type of counterfactual feedback provided. 
When provided by the human, we use the triple for an additional update of $H$, combined with the standard TAMER update $\langle \mathit{f}, s, a \rangle$. The full procedure with all steps is also described in Algorithm \ref{algo:tamer}.

\subsection{Using the contrastive nature of counterfactuals}
\label{sec:Using the contrastive nature of counterfactuals}
Besides using counterfactuals as additional data points, we will also leverage another essential property of counterfactuals: their natural contrast between the \textit{fact} and the \textit{foil}.
This contrast allows us to introduce an additional term with contrastive constraints in the loss function to distance the activations of each state-action pair from each other.

In Equation \ref{eq:constrastive_loss}, our contrastive loss is shown, where $E(s, a)$ is the output (or embedding) of the last-hidden layer, which is common to the counterfactual and feedback pair (depending on the respective architecture for the H-Model). This loss term should speed up the learning process due to the additional information contained in the gradients.

\begin{align}
&\label{eq:constrastive_loss}L_{cos}(s_{t-1}, a_{t-1}, s_{\mathit{cf}}, a_{\mathit{cf}}) = {}\\
&\nonumber\mspace{100mu}max(0, cos(E(s_{t-1}, a_{t-1}), E(s_{\mathit{cf}}, a_{\mathit{cf}}))
\end{align}

With this contrastive term, the complete loss (see Equation \ref{eq:full_loss}) can be built as a simple combination of all three loss terms. The \textit{$L_{normal}$} and the \textit{$L_{\mathit{cf}}$} could be any loss classes, in our case, they both refer to a L2-loss.
\begin{align}
&\label{eq:full_loss}L(s_{t-1}, a_{t-1}, \mathit{f}, s_{\mathit{cf}}, a_{\mathit{cf}}, \mathit{f}_{\mathit{cf}}) = {}\\ 
&\nonumber\mspace{100mu}L_{normal}(s_{t-1}, a_{t-1}, \mathit{f}) + {}\\
&\nonumber\mspace{100mu}L_{\mathit{cf}}(s_{\mathit{cf}}, a_{\mathit{cf}}, f_{\mathit{cf}}) + {}\\
&\nonumber\mspace{100mu}L_{cos}(s_{t-1}, a_{t-1}, s_{\mathit{cf}}, a_{\mathit{cf}})
\end{align}


\section{Evaluation}


For our evaluation, we will use TAMER as the baseline, and compare it to the proposed additions of counterfactuals states (TAMER+CFS) and counterfactual actions (TAMER+CFA). 
First, we evaluate the performance of each variant in different environments and show that our changes lead to faster learning.
Afterwards, we introduce changes in the feedback frequency and in the quality of the feedback to show that the improvement with counterfactual feedback also holds under more realistic feedback conditions.
Finally, we perform some simple ablations to evaluate our choices regarding the direction of counterfactuals and their benefit over random samples.


\subsection{Environment Setup \& Measurements}
\label{sec:Environment Setup}

We run the evaluations in multiple environments (CartPole, LunarLander, Minigid \footnote{https://github.com/maximecb/gym-minigrid}). We changed the MiniGrid environment, so that the goal position is random (instead of a static goal at the bottom-right corner.)  We selected an architecture similar to the one used in DeepTAMER \cite{Warnell2018DeepTI}.

To evaluate the performance of the learning agent, we use the environmental reward as a measure of success. This reward is only used to evaluate the agent and is never seen or used in training. The agent fully learns from human feedback. In the following sections, we will refer to this reward for the evaluation of the algorithms. 
To track the learning progress, at fixed steps, we run an evaluation (without learning) in 10 different (fixed) seeds of the same environment. The total reward of each run is collected and then presented as a mean. This allows us to track the learning of an agent throughout its training. Since the learning of HRL can be highly unstable, we perform multiple training runs per algorithm/environment with different random seeds to initialize the agent and the environment. To allow for easier comparisons, we normalize the rewards at the end of each episode so that a (perfect) solution results in the total reward of 1, while random agents get a 0 reward (on average). 
Since we want to enable faster learning and therefore want to look at the whole training process (instead of the performance after the training), we selected the optimality gap (the area between the current performance and a perfect performance) as a metric for the evaluation. This metric enables us to evaluate the performance of our variants in a statistically sound way, without the pitfalls of showing large amounts of metrics in large tables or interpreting graphs, as discussed by Agarwal \etal \cite{Agarwal2021DeepRL}.

With our scores already normalized, an utterly random agent would receive an optimality gap of 1, while a perfect agent would produce a 0. Additionally, we compute 95\% confidence intervals through bootstrapping (as proposed by \cite{Agarwal2021DeepRL}) in each plot, indicated by the shaded area.


\subsection{Human-Feedback Oracle}
\label{sec:Human-Feedback Oracle}
While HRL aims to incorporate human feedback into the training loop, it can be costly to collect human feedback.
This cost can hinder the development and evaluation of HRL algorithms since evaluations are often performed with only a few humans, leading to false interpretations due to statistical randomness. 
In this work, we focus purely on the improvements of the feedback from a learning standpoint. Therefore, we use a synthetic oracle to imitate human feedback (like other prior work, e.g., by Christiano \etal \cite{christiano2017deep}). To achieve this, we trained an agent for the oracle with proximal policy optimization (PPO) \cite{ppo} algorithm on every environment for sufficiently many steps to solve the tasks. We then use this policy to give dynamically calculated (counterfactual) feedback for the actual learning algorithm. 

The handling of inconsistent and sub-optimal feedback is one of the major challenges in HRL. Our evaluation accounts for inconsistency and sub-optimality by introducing two hyper-parameters. With the ``feedback frequency'' parameter, we control how often the oracle gives feedback towards the agent to archive inconsistent feedback. With the ``feedback quality'' parameter, we allow the oracle to give random feedback, which results in lower-quality feedback. 

\begin{figure*}[h!]
     \centering

        \includegraphics[width=.99\textwidth]{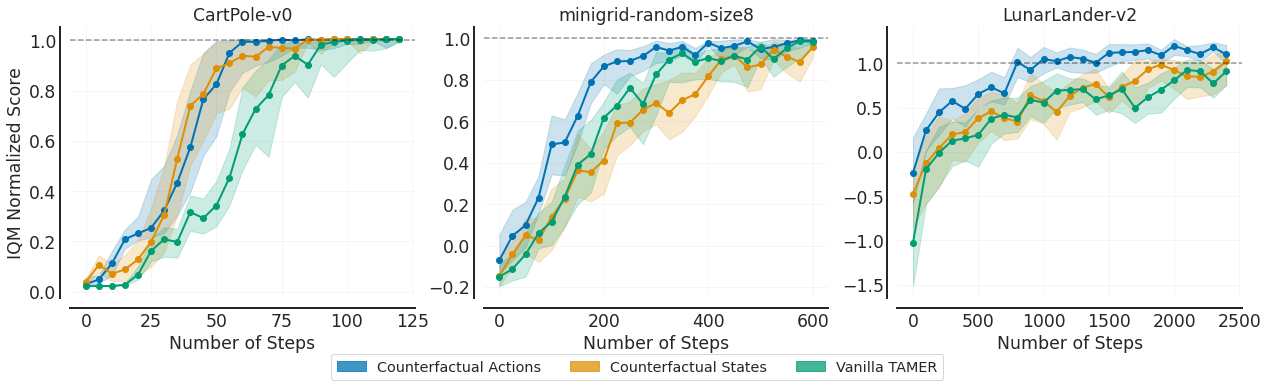}

        \caption{Interquartile mean reward of each feedback method on the evaluation set throughout training. Bold line is the IQM reward of random seeds (n=25), area is the 95\% confidence interval.}
             \label{fig:differentshapes}
\label{fig:performance}
\end{figure*}

\subsection{Results}

As shown in Figure \ref{fig:performance}, TAMER extended with counterfactuals leads to faster learning than the original TAMER. The main advantage is at the early stages of training, where learning is significantly faster than the original TAMER. At later stages of the training, the difference shrinks when all variants (vanilla TAMER and our extensions) reach the maximum reward plateau (with enough training steps). 

\begin{figure*}[t]
    \centering
    \includegraphics[width=.99\textwidth]{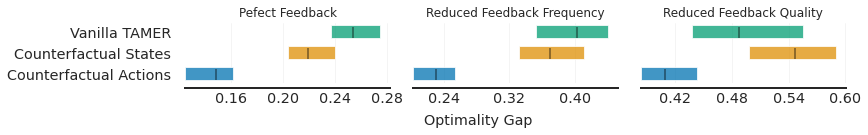}
    \caption{Optimality Gap with confidence intervals on all tasks. (Lower means better)}
    \label{fig:optimality_gap_normal}
\end{figure*}

To highlight the faster learning, we measure the absolute performance of each variant at regular intervals for each environment and calculate the optimality gap with these measurements. The results can be seen in Figure \ref{fig:optimality_gap_normal}, which shows that both of our methods learns faster than vanilla TAMER. Counterfactuals actions perform better than counterfactuals states, since counterfactual actions outperform vanilla TAMER, while counterfactual states struggle in some environments.

\textbf{Reduced feedback frequency.} 
Perfect feedback rates are not a realistic assumption for HRL. Hence, we investigate our extensions under more realistic conditions: we reduce the frequency in which the oracle gives feedback. This mimics one essential characteristics of HRL, namely that the feedback is not deterministic and can be inconsistent and sporadic \cite{HumanFeedbackMotivation}. To achieve this, we change the feedback frequency of our oracle and evaluate our extension under this changed condition.
As shown in Figure \ref{fig:optimality_gap_normal}, the results show that even with reduced feedback rates, the general pattern of an increase in learning with the inclusion of counterfactuals is similar to the results with a perfect feedback rate. We noted that some environments suffer more from the reduced rate of feedback.

\textbf{Reduced feedback quality.} Another essential characteristic of human feedback is that it does not have to be optimal at every point in time. Next, we evaluate the learning behavior for lower-quality feedback. We try to approximate these with the synthetic oracle (see Section \ref{sec:Human-Feedback Oracle}), for which set the ``feedback quality'' parameter to 0.75. 
The results show that counterfactual actions (CFA) still work with lower quality and outperform the baseline. Naturally, the learning is slower with less optimal feedback since the ``signal strength'' is lower. It can be noted that counterfactual states struggle to perform on a same level than the baseline.

\textbf{Downward Counterfactuals}
The usage of upward counterfactuals was motivated by the performance differences lemma (see section \ref{sec:The convergence of Reinforcement Learning algorithms} and the preference of humans. We include a downward counterfactual condition in the evaluation to evaluate whether the upward property is essential.  

\begin{figure}[t]
    \centering
    \includegraphics[width=0.5\textwidth]{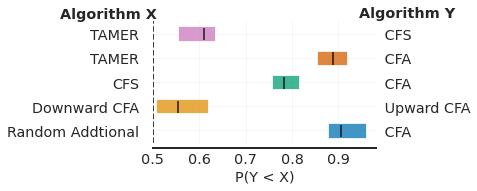}
    \caption{Placeholder Probability of Improvement, calculated on the optimality gap, so a improvements means a lower score $P( Y < X)$,  for various ablations}
    \label{fig:poi_ablation}
\end{figure}

The results can be seen in Figure \ref{fig:poi_ablation}, for which we calculate on each comparison the probability that one variant leads to a better optimality gap than the other. 
Using downward counterfactuals still improves the learning compared to vanilla TAMER, but upward-directed counterfactuals lead to faster learning than downward counterfactuals. 

\textbf{Random Samples}
Since we treat counterfactual explanations as additional samples, one could argue that, by counterfactual explanations, users just submit more samples to the learning agent, and therefore the learning process converges faster.
To control for this possibility, we evaluate a Random condition. In this condition, the oracle provides a random state/action tuple, which is correct but totally unrelated to the current state. In this case we disable the contrastive loss, since there is no guarantee for a contrast between the current and the random tuple. Figure \ref{fig:poi_ablation} shows that indeed the counterfactual samples perform superior to the random samples.
%

\subsection{Discussion}

\textbf{Faster learning.} We extended TAMER to allow the human to give additional counterfactual feedback and evaluated our extension in various environments and under multiple conditions. The results indicate that the inclusion of counterfactuals increases the learning in every condition. Thus, counterfactual feedback is handled more label efficient than simple binary feedback or random explanations. This behavior confirms the expectation stated in section \ref{sec:The convergence of Reinforcement Learning algorithms}: In the early stages of training, a large difference between the agent and the expert (human) policy exists. Therefore, introducing a single example from the expert policy can have a significant impact. In later stages of the training, when the difference shrinks, the impact of additional expert samples becomes lower. A limiting side-effect of the learning is that the agent sees less and less upward counterfactual in the learning process since we only give counterfactual feedback in negative cases, which occur fewer in later stages of the training. 

\textbf{Less and lower quality feedback.} In cases where the feedback is of lower quality (which also makes the counterfactual feedback of lower quality), our counterfactual actions still proves to increase the rate of learning. Our counterfactual states struggle with reduced feedback quality. This is due to our construction of the counterfactual feedback. Usually, the state space is much larger than the action space. Since our lower quality feedback for counterfactuals randomly samples the respective state or action space (for either counterfactual states or actions), this leads to a lower signal strength in counterfactual states. Further research could be done to determine acceptable lower bounds for the quality of human feedback.


\textbf{Counterfactual states vs.\ actions.} Both extensions, counterfactual actions and counterfactual states, increase the rate of learning, with counterfactual actions (CFA) providing better results than counterfactual states (CFS). We reason this is because counterfactual actions provide more immediate useful feedback than counterfactual states feedback. This, in turn, can be explained since with counterfactual actions, the agent already knows how to reach state $s$, and it can use the counterfactual action $a_{\mathit{cf}}$ the next time the agent reaches it.

\textbf{Upward Counterfactuals \& Random Feedback} Our main focus in this work was on upward counterfactuals. Since they come directly from the expert policy (see \ref{sec:The convergence of Reinforcement Learning algorithms}), we expected them to perform better than downward directed counterfactuals. Additionally, as discussed,  upward counterfactuals are preferred by humans \cite{CounterfactualThinking,CounterfactualThinking2}. Nevertheless, we evaluated the impact of downward counterfactuals. To no surprise, they perform not as well  as upward counterfactuals but still benefit the learning. Similarly, we evaluated the case when additional random feedback samples are given, which performed even worse. This highlights that using explanation concepts can provide more natural interfaces towards the human and provide more benefits than just simple samples. In this work we have shown that counterfactuals, especially upward-directed ones, provide a clear benefit from a learning perspective while giving the human more freedom.  We think these results can be seen as motivation to investigate more explanations concepts.

\textbf{Realistic oracles.} With our implementation of a synthetic oracle, we took a step to make the synthetic oracle more realistic due to the randomness in feedback frequency and the reduced optimality of the feedback. This is far from a perfect simulation of human feedback but the oracle also allowed us to run experiments with a large sample size in multiple environments with ablations. We believe that our reduced frequency and optimality are realistic enough to ensure that the results of our extensions will still hold in practice. 

\textbf{Interfaces for counterfactual feedback.} Throughout our work, we have shown that counterfactuals are beneficial for learning. However, it is still an open question how humans would actually give this kind of feedback to the agents. While the handling of more complex feedback is definitely a complex problem, one should note there are many instances when humans naturally want to provide more complex feedback. We argue that, depending on the use case, humans can be enabled to give complex feedback through better interfaces. For example, an NLP-based parsing of utterances could extract counterfactuals from humans. Alternatively, for environments like MiniGrid, one could provide humans with a graphical interface, where they can not only give feedback but also change the state (by dragging objects around) to produce counterfactual states.

\textbf{Feedback Cost} 
Our results show that the inclusion of counterfactuals leads to faster learning in terms of fewer steps in the environment needed before it is `solved`. This also leads to fewer feedback interactions from humans. While this is a preferable result, counterfactual feedback could also lead to additional mental workload by the human. This  load was not considered in our evaluation. However, studies suggest that in many situations humans naturally want to enrich their feedback with additional explanations (see \cite{HumanExperienceIRL,UserFeedbackMachineLearning}), without much thinking. Our method gives  humans more flexibility in their feedback and allows them to harness it. 
We believe that the additional mental load depends on the environment and the feedback interface and has to be evaluated on a case-by-case basis. Additionally, we think it could benefit the user experience if the user could express his feedback differently.

\section*{Conclusions}
We have shown, with the help of synthetic oracle, that the inclusion of upward counterfactual explanations in TAMER can significantly improve the speed of learning as compared to simple binary feedback, random feedback, or downward counterfactual feedback. Both approaches, counterfactuals based on states and counterfactuals based on actions, show these improvements. 
%
Overall counterfactual actions delivered the lowest optimality gap, speeding up the learning the most, while counterfactual states still provide benefits over TAMER without counterfactuals. We tried to circumvent the limitations of the synthetic oracle with experiments in which the quality of the oracle had been reduced. The results of these experiments showed that the improvements should also hold under more realistic conditions. Nevertheless for definite results further studies with human trainers are needed.
Future work will investigate how users actually use these enhanced feedback interfaces and how costly for the human they are. 



\bibliographystyle{IEEEtran}
\bibliography{IEEEabrv,sources}
\end{document}